\documentclass[10pt, a4paper]{article}

\usepackage[final]{lrec-coling2024} % this is the new style

\usepackage{algorithm}
\usepackage{graphics}
\usepackage{algpseudocode}
\usepackage{amsmath}
\usepackage{enumitem}
\usepackage{mathtools}
\usepackage{multirow}
\usepackage{graphicx}
\usepackage{booktabs}       % professional-quality tables
\usepackage{amsmath}
\usepackage{amssymb}% http://ctan.org/pkg/amssymb
\usepackage{pifont}% http://ctan.org/pkg/pifont
\newcommand{\cmark}{\ding{51}}%
\newcommand{\xmark}{\ding{55}}%

\title{Federated Foundation Models: Privacy-Preserving and Collaborative Learning for Large Models}

\name{ Sixing Yu$^{1}$, \space J. Pablo Muñoz$^{2}$, \space  Ali Jannesari$^{1}$} 

\address{ $^{1}$Iowa State University\\ 
  $^{2}$Intel Labs\\
  \texttt{\{yusx, jannesar\}@iastate.edu, pablo.munoz@intel.com}\\}

\abstract{
Foundation Models (FMs), such as LLaMA, BERT, GPT, ViT, and CLIP, have demonstrated remarkable success in a wide range of applications, driven by their ability to leverage vast amounts of data for pre-training. However, optimizing FMs often requires access to sensitive data, raising privacy concerns and limiting their applicability in many domains.
In this paper, we propose the Federated Foundation Models (FFMs) paradigm, which combines the benefits of FMs and Federated Learning (FL) to enable privacy-preserving and collaborative learning across multiple end-users.
We discuss the potential benefits and challenges of integrating FL into the lifespan of FMs, covering pre-training, fine-tuning, and application.
We further outline potential future research avenues in FFM, including FFM pre-training, FFM fine-tuning, and federated prompt tuning, which allow the development of more personalized and context-aware models while ensuring data privacy.
Moreover, we explore the possibility of continual/lifelong learning in FFMs, as increased computational power at the edge may unlock the potential for optimizing FMs using newly generated private data close to the data source.
The proposed FFM concepts offer a flexible and scalable framework for training large language models in a privacy-preserving manner, setting the stage for subsequent advancements in both FM training and federated learning. \\ \newline  }

\begin{document}
\maketitleabstract
\section{Introduction}
In recent years, Foundation Models (FMs) such as BERT~\cite{kenton2019bert}, GPT~\cite{brown2020gpt3,radford2019gpt2}, Llama~\cite{touvron2023llama,touvron2023llama2}, ViT~\cite{dosovitskiy2020vit}, and CLIP~\cite{radford2021clip} have significantly advanced the field of artificial intelligence, showcasing impressive performance across a wide range of tasks and domains. However, the optimization of increasingly complex FMs heavily depends on the collections of massive datasets, which introduces concerns regarding training data scarcity, computational resources, privacy, and ethical considerations.
Simultaneously, the prevalent trend of advancement in edge technologies generates a vast amount of decentralized data, creating potential resources for further optimizing and specializing FMs. Nevertheless, due to privacy concerns, this private data is rarely leveraged for FM optimizations. In light of this, Federated Learning (FL)~\cite{mcmahan2017fedavg} has emerged as a pioneering approach for decentralized and privacy-preserving machine learning, allowing models to learn from distributed private data sources without directly accessing the raw data.
\begin{figure*}
  \centering
  \includegraphics[width=.6\linewidth]{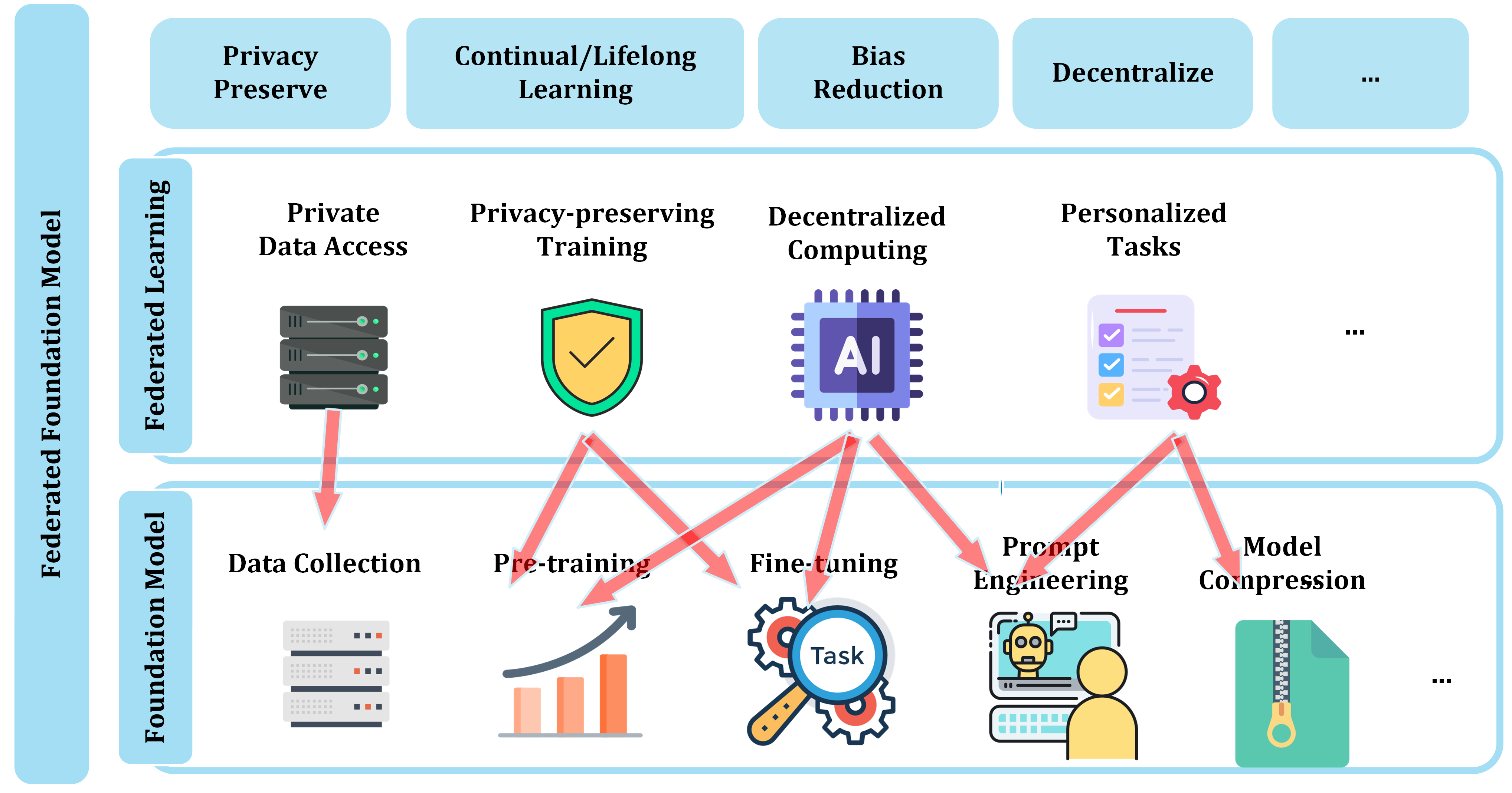}
  \caption{Federated Foundation Model: Integrating federated learning into the lifespan of foundation models, facilitating privacy-preserving, scalable, lifelong learning, robustness, and decentralized FMs.}
  \label{fig:FFM overview}
\end{figure*}

The intersection of these two domains presents a unique opportunity to unlock new possibilities in AI research and to address critical challenges in AI model development and real-world applications.
Hence, we propose the concept of Federated Foundation Models (FFMs), a novel paradigm that integrates FL into the lifespan of FMs. This integration addresses the challenges mentioned above related to data scarcity, computational resources, privacy, and ethical considerations while facilitating privacy-preserving and collaborative learning across multiple end-users. As advancements in edge computing enable the optimization of FMs using FL, we further explore the possibility of continual/lifelong learning for FMs in FFMs.
We also discuss the potential benefits and challenges of integrating FL into different stages of the FMs' lifespan, including pre-training, fine-tuning, and application, and provide potential research directions for FFM tasks such as FFM Pre-training, FFM Fine-tuning, and Federated Prompt Tuning. These tasks promote the development of personalized and context-aware models while maintaining data privacy.

In summary, this paper offers a comprehensive examination of the prospective of FFMs, proposing a flexible and scalable framework for training large models in a privacy-preserving manner. We believe our work contributes to paving the way for future advancements in both FMs and FL, fostering the development of more secure and adaptable large models and FL algorithms that cater to a wide range of applications.
%In summation, this paper presents a thorough exploration of Federated Foundation Models, proposing a flexible and scalable framework for training large models in a privacy-preserving manner. We posit that our work lays the groundwork for future advancements in both FMs and FL, promoting the creation of more secure and adaptable large models and FL algorithms suited for a diverse array of applications.

\section{Background}
\subsection{Federated Learning}
As concerns about user data privacy grow, there is an increasing need for AI models to be trained on decentralized data without sharing private information between clients. Federated Learning (FL) has emerged as a solution to this problem, offering a distributed and privacy-preserving machine learning approach that enables training on decentralized data without compromising data privacy~\cite{mcmahan2017fedavg}.

In FL, raw data remains on local clients, ensuring data privacy and security while also enabling collaborative learning across multiple clients. The FL process involves local model training, model aggregation algorithm, and global model updates. Throughout this process, clients only share model updates, such as weights and gradients, asynchronously, reducing bandwidth requirements and minimizing the risk of data leaks and breaches.
A typical FL algorithm is FedAvg~\cite{mcmahan2017fedavg}, which demonstrates the FL process (see Algorithm~\ref{algo:fedavg}). The privacy-preserving nature of FL has led to its widespread adoption in various applications, particularly in privacy-sensitive domains like healthcare.

\begin{algorithm}
\caption{Federated Learning Process (FedAvg)}
\label{algo:fedavg}
\begin{algorithmic}[1]
\State \textbf{Input:} Global AI model $w_0$, clients $S$, communication rounds $T$
\For{$t = 1, 2, \ldots, T$}
\State Server deploys global model $w_{t-1}$ to clients $\in S$
\For{each client $k \in S$}
\State Client $k$ optimizes $w_{t-1}$ on local data, producing $w_{t}^k$
\EndFor
\State Select a subset of clients $S_t$ to communicate with the server
\For{each client $k \in S_t$}
\State Client $k$ sends local model update $\Delta w_t^k = w_t^k - w_{t-1}$ to the server
\EndFor
\State Server aggregates local updates and computes the new global model:
\begin{equation*}
w_t = w_{t-1} + \eta_t \sum_{k \in S_t} n_k \Delta w_t^k
\end{equation*}
\EndFor
\end{algorithmic}
\end{algorithm}

However, FL still faces challenges related to heterogeneous data distribution. Data may be non-independent and identically distributed (non-IID) across clients, leading to poor model convergence and performance. Recent work in FL has focused on improving gradient descent to stabilize training~\cite{liu2020momentum_fl,karimireddy2020scaffold,yu2021feddp}; personalizing model weights to enhance performance on downstream tasks~\cite{deng2020adaptive_pfl,tan2022personalizedfl,yu2022rafl,yu2022spatl}; and employing model compression techniques like knowledge distillation, dynamic dropout, and adaptive pruning to reduce overfitting on non-IID datasets and improve communication efficiency ~\cite{jiang2022pruneFL,yu2021feddp,lin2020feddf,yu2021feddp,lin2020feddf,yu2022spatl,yu2022kdfl,nguyen2024flkd}.
Despite these advances, there remains a gap between traditional model training and FL, particularly in terms of performance when dealing with heterogeneous data distributions.

\subsection{Foundation Models}
Foundation Models (FMs), such as the GPT family~\cite{brown2020gpt3,radford2019gpt2}, ViT~\cite{dosovitskiy2020vit}, CLIP~\cite{radford2021clip}, and BERT~\cite{kenton2019bert}, have become a driving force in AI, serving as the basis for various downstream tasks. These models are trained on massive datasets and demonstrate remarkable capabilities across multiple domains.
The lifespan of FMs typically includes pre-training, fine-tuning, and application. Pre-training involves unsupervised or self-supervised learning on large-scale datasets, while fine-tuning adapts the models to specialized tasks. For example, GPT~\cite{brown2020gpt3,radford2019gpt2,openai2023gpt4} models learn grammar, syntax, and semantics during pre-training, enabling them to be easily fine-tuned for tasks such as text classification, sentiment analysis, translation, and summarization. Parameter-efficient fine-tuning (PEFT) methods, e.g., low-rank adapters (LoRA) \cite{hu2022lora}, have been proposed to reduce the memory and compute requirements during the fine-tuning of these large models. Recently, neural architecture search (NAS) techniques have been employed to discover high-performing configurations of these adapters \cite{lonas2024, shears2024}.

In the application stage, FMs show extraordinary adaptability to downstream tasks using zero-shot learning. Prompt Engineering, an emerging research area, explores this potential by optimizing the interaction between users and FMs through carefully crafted prompts, thereby improving performance on downstream tasks. Various methods for prompt engineering have been proposed, including prompt templates~\cite{wei2021prompt}, prompt tuning and instruction tuning~\cite{wei2021prompt}~\cite{lester2021prompt_tune,han2022ptr_prompt_tune}, automated prompt generating~\cite{zhou2022automated_prompt,sanh2021multitask_automated_prompt}, and in-context learning~\cite{min2021icl_metaicl,min2022icl_rethinking,rubin2021icl_retrive_prompt,liu2021icl_makes}. These approaches enable FMs to learn from examples or instructions supplied as part of the input without the need for explicit fine-tuning or labeled examples.

In summary, the combination of Federated Learning and Foundation Models offers great opportunities to revolutionize the AI landscape by leveraging the strengths of both paradigms. This intersection opens up numerous research directions and applications in areas such as personalized recommendations, natural language understanding, healthcare, finance, and more. As AI researchers continue to explore Federated Foundation Models, we expect to see innovative solutions and breakthroughs that lead to more robust, efficient, and ethical AI systems serving the needs of individuals and society.

% \usepackage{multirow}
% \usepackage{graphicx}
\begin{table*}[]
\centering
\caption{Comparison of the Federated Foundation Model with Traditional FM Optimization}
\label{tab:comparison}
\resizebox{.8\textwidth}{!}{%
\begin{tabular}{ccccc}
\hline
                                                                  & \multicolumn{2}{c}{\textbf{\begin{tabular}[c]{@{}c@{}}Federated \\ Foundation Model\end{tabular}}}                             & \multicolumn{2}{c}{\textbf{\begin{tabular}[c]{@{}c@{}}Traditional \\ FM Optimization\end{tabular}}}                                  \\ \hline
\multirow{2}{*}{Data Privacy}                                     & \multirow{2}{*}{Privacy-preserve}                                                                  & \multirow{2}{*}{\cmark} & \multirow{2}{*}{Centralized Data Collection}                                                             & \multirow{2}{*}{\xmark} \\
                                                                  &                                                                                                    &                           &                                                                                                          &                           \\ \hline
\begin{tabular}[c]{@{}c@{}}Communication \\ Overhead\end{tabular} & Communicate Model Updates                                                                          & \cmark                  & Communicate Data to Central Server                                                                       & \xmark                  \\ \hline
\begin{tabular}[c]{@{}c@{}}Model \\ Performance\end{tabular}      & Diverse Data Improvement                                                                           & \cmark                  & Lacks Diversity                                                                                          & \xmark                  \\ \hline
\begin{tabular}[c]{@{}c@{}}Resource \\ Distribution\end{tabular}  & Distributed Across Devices                                                                         & \cmark                  & Centralized                                                                                              & \xmark                  \\ \hline
\begin{tabular}[c]{@{}c@{}}Data \\ Efficiency\end{tabular}        & Better with data diversity                                                                         & \cmark                  & \begin{tabular}[c]{@{}c@{}}Requires more data for \\ similar performance\end{tabular}                    & \xmark                  \\ \hline
\multirow{2}{*}{Latency}                                          & \multirow{2}{*}{Distributed Computation}                                                           & \multirow{2}{*}{\xmark} & \multirow{2}{*}{Lower with Centralized Computation}                                                      & \multirow{2}{*}{\cmark} \\
                                                                  &                                                                                                    &                           &                                                                                                          &                           \\ \hline
\begin{tabular}[c]{@{}c@{}}System \\ Complexity\end{tabular}      & Distributed Coordination                                                                           & \xmark                  & Centrally Managed                                                                                        & \cmark                  \\ \hline
\multirow{2}{*}{Scalability}                                      & \multirow{2}{*}{Scalable to Many Clients}                                                          & \multirow{2}{*}{\cmark} & \multirow{2}{*}{Unscalable with Large Datasets}                                                          & \multirow{2}{*}{\xmark} \\
                                                                  &                                                                                                    &                           &                                                                                                          &                           \\ \hline
\multirow{2}{*}{Consistency}                                      & \multirow{2}{*}{\begin{tabular}[c]{@{}c@{}}Weakly Connected\\ Collaborative Learning\end{tabular}} & \multirow{2}{*}{\xmark} & \multirow{2}{*}{\begin{tabular}[c]{@{}c@{}}Consistent Updates in \\ Controlled Environment\end{tabular}} & \multirow{2}{*}{\cmark} \\
                                                                  &                                                                                                    &                           &                                                                                                          &                           \\ \hline
\begin{tabular}[c]{@{}c@{}}Ease of\\ Deployment\end{tabular}      & Challenging                                                                                        & \xmark                  & Easier                                                                                                   & \cmark                  \\ \hline
\end{tabular}%
}
\end{table*}
\section{Motivation for Federated Foundation Models}
\label{sec:prospective}

In this section, we discuss the various challenges that motivate the development of Federated Foundation Models (FFMs), covering aspects such as data privacy, model performance, communication cost, scalability, deployment, personalization and real-time adaptation, and bias reduction. As shown in Figure~\ref{fig:FFM overview}, These existing challenges highlight the potential advantages of combining Foundation Models (FMs) and Federated Learning (FL) for a wide range of applications and scenarios.

\noindent\textbf{Data privacy.}
The widespread deployment of AI in society generates vast amounts of data (e.g., images collected by cameras in smartphone applications, prompt dialog produced by users), presenting potential resources for optimizing and specializing FMs.
However, privacy concerns have limited the use of private data for FM optimization. 
FFMs offer significant improvements in data privacy by incorporating FL, enabling FM optimization on private data. 
By optimizing FM tasks~(e.g., pre-training, fine-tuning, and prompt tuning) on local data without sharing raw information, FFMs comply with data protection regulations and preserve user privacy. This approach is particularly beneficial when sensitive data, such as medical records or personal communications, must be used to improve model performance without compromising confidentiality.

\noindent\textbf{Model performance.}
Combining FMs and FL provides benefits to FMs, boosting their performance. FMs gain access to a broader range of data for optimization tasks such as fine-tuning, prompt tuning, and pre-training. This expanded data access enables the development of more accurate and efficient AI systems better suited for users in diverse scenarios. This combination benefits FL, as well. FL can overcome challenges associated with Non-IID~(Non-Identical Independent Distributed) and biased data~\cite{zhao2018noniid} by leveraging the advanced capabilities of FMs, leading to improved performance across different tasks and domains.

\noindent\textbf{Cost.}
FFMs reduce communication costs by sharing only model updates between devices and the central server, significantly saving bandwidth and communication costs for transmitting raw data. Additionally, FFMs can potentially reduce the labor cost associated with collecting and managing data in a central location, as data is generated and used locally at edge devices. This efficiency makes FFMs a more practical and cost-effective solution for training and deploying FMs.

\noindent\textbf{Scalability.}
Current FMs, especially large language models, often face scalability limitations due to limited computational power at the edge. Many FMs are run centrally and provide API access for users, which can lead to capacity constraints and API congestion. In the near future, advancements in computational power may enable FMs to run locally on edge devices. FL's scalable nature makes it an ideal framework for combining with FMs, accommodating numerous devices with varying computational capabilities. By integrating FL principles, FMs can leverage advancements in computational power, becoming more scalable and enabling broader deployment and improved performance across various tasks and domains.

\noindent\textbf{Deployment.}
FFMs offer potential advantages in deployment, particularly in reducing latency and enhancing user experience. Running FMs centrally with API access for users can result in latency issues due to network communication between the user's device and the central server hosting the model. In contrast, FFMs can be deployed and run locally on edge devices, potentially reducing latency by eliminating network communication. This allows for faster response times and a more seamless user experience when interacting with the model. However, available computational resources on edge devices must be considered when deploying FMs locally. As discussed in the Scalability section, advancements in computational power will be crucial for enabling local deployment on a wide range of devices, ensuring efficient and effective performance across various tasks and domains.

\noindent\textbf{Personalization and real-time adaptation.}
FFMs facilitate a high degree of personalization by leveraging the decentralized nature of FL. By training on diverse, user-generated data, FMs can be tailored to individual preferences and requirements, offering more personalized and context-aware solutions across various tasks and domains.
A key advantage of FFMs is their ability to adapt in real-time as new personalized data becomes available from edge devices. This continuous learning capability ensures that the models remain up-to-date with users' evolving needs and preferences, further enhancing their personalization.
The focus on personalization in FFMs leads to improved performance and greater user satisfaction. By providing AI solutions that dynamically adapt to user-specific needs, FFMs enable more effective and engaging user experiences across a wide range of applications and domains.

\noindent\textbf{Bias reduction.}
FFMs contribute to bias reduction in AI systems by incorporating diverse data from decentralized sources, resulting in more inclusive and fair AI solutions. The models learn from various users, increasing their awareness of the nuances and complexities of real-world scenarios, and leading to more informed and less biased decisions across tasks and domains.
Additionally, the privacy-preserving nature of FL encourages more users to participate in the training process, further diversifying the data and knowledge incorporated into FMs. This results in models better equipped to handle and minimize biases, providing fairer and more equitable AI solutions for all users.

\noindent\textbf{Continual/Lifelong learning.} 
FMs combined with FL provide an ideal platform for continual lifelong learning. This combination facilitates the continuous adaptation and improvement of models by harnessing decentralized and diverse data sources, leading to more versatile and effective AI systems. As advancements in edge computing power become more prevalent, the realization of continual lifelong learning in FMs will soon be within reach. This progress will enable AI models to learn and grow throughout their lifespan, unlocking new possibilities for AI research and practical applications in various domains. By embracing continual lifelong learning, FFMs can help create more adaptive, efficient, and personalized AI systems that can dynamically adjust to user-specific needs and preferences, ultimately benefiting users from all walks of life.

In summary, as detailed in Table~\ref{tab:comparison}, 
our proposed FFM presents several advantages over traditional FM optimization. Despite introducing certain challenges, FFMs exhibit significant promise in enhancing data privacy, reducing communication overhead, improving model performance, optimizing resource distribution, increasing data efficiency, and providing better scalability. FFMs represent a robust approach to address many challenges and limitations associated with traditional, centralized machine learning. By incorporating Federated Learning (FL) into FM optimization, we are poised to engender more efficient, personalized, and privacy-conscious AI systems. This advancement heralds a new era in AI research and application, potentially making AI more equitable and advantageous for a diverse array of users. The integration of FL not only fortifies the foundational aspects of machine learning but also democratizes AI, thereby extending its benefits across a broader societal spectrum.

\begin{figure*}
  \centering
  \includegraphics[width=\linewidth]{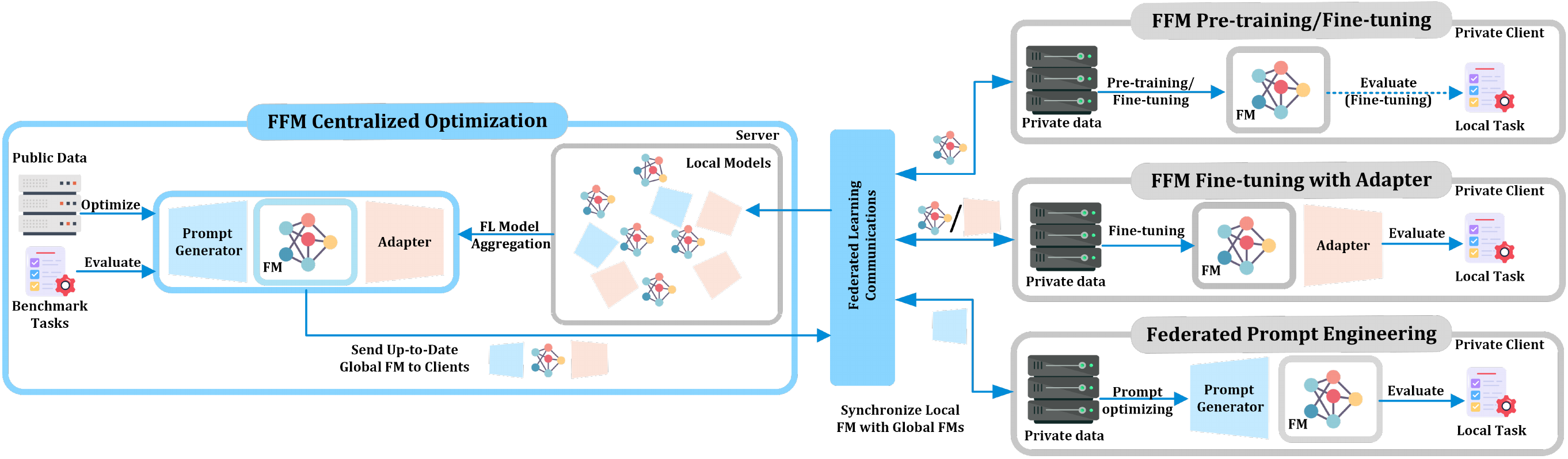}
  \caption{Federated Foundation Model tasks: The FFM centralized optimization process aggregates local models and updates them using public data. Private clients download up-to-date global model parameters from the server, optimize the FM locally on their tasks, and send the optimized model back to the server. }
  \label{fig:FFM_tasks}
\end{figure*}

\section{Federated Foundation Model: Prospective and Future Research}
In this section, we discuss potential future research directions and general challenges related to FFMs, covering but not limited:
\begin{itemize}
\item Federated foundation model pre-training
\item Federated foundation model fine-tuning
\item Federated prompt tuning
% \item Federated foundation model deployment
\item Federated continual (lifelong) learning
\item Federated retrieval augmented generation
\item General challenges
\item Other future research directions
\end{itemize}
% We examine the unique characteristics and requirements of these tasks, highlighting the opportunities and challenges that arise when leveraging FFMs to tackle real-world problems. Our goal is to establish a solid foundation for understanding the scope and potential of this new paradigm, paving the way for future research and development. 
We scrutinize the distinct characteristics and prerequisites of these tasks, spotlighting the opportunities and hurdles encountered when employing FFMs to address real-world issues. Our aim is to build a robust foundation for comprehending the breadth and potential of this emerging paradigm, thereby fostering further research and development.
As mentioned in Section~\ref{sec:prospective}, some tasks may not be feasible until computational power at the edge advances further.
% We scrutinize the distinct characteristics and prerequisites of these tasks, spotlighting the opportunities and hurdles encountered when employing FFMs to address real-world issues. Our aim is to build a robust foundation for comprehending the breadth and potential of this emerging paradigm, thereby fostering further research and development. As noted in Section~\ref{sec:prospective}, certain tasks may remain unattainable until edge computational prowess progresses.

\begin{algorithm}
\caption{General FFM Optimization process}
\label{algo:ffm}
\begin{algorithmic}[1]
\State \textbf{Input:} Global AI model $w_0$, clients $S$, communication rounds $T$
\State Server initialize global model $w_0$
\For{$t = 1, 2, \ldots, T$}
\If {Public data available} 
\State{Server optimize $w_{t-1}$ on public data }
\EndIf
\State Server send global model $w_{t-1}$ to participate clients $\in S$
\For{each client $k \in S$} \textbf{ in parallel}
\State Client $k$ optimizes $w_{t-1}$ on local data, producing $w_{t}^k$
\EndFor
\State Select a subset of clients $S_t$ to communicate with the server
\For{each client $k \in S_t$}
\State Client $k$ sends local model update $\Delta w_t^k = w_t^k - w_{t-1}$ to the server
\EndFor
\State Server aggregates local updates and computes the new global model:
\begin{equation*}
w_t = w_{t-1} + \eta_t \sum_{k \in S_t} n_k \Delta w_t^k
\end{equation*}
\EndFor
\end{algorithmic}
\end{algorithm}

\subsection{Pre-training of Federated Foundation Models}
\label{sec:pt_ffm}

% \subsubsection{Task Definition}
\textbf{Motivation:}
The motivation behind Federated Foundation Model (FFM) pre-training is to enhance traditional Foundation Model (FM) pre-training methodologies, harnessing Federated Learning's (FL) capability to utilize private data to improve model generalization while preserving data privacy. Introducing FL to FM lifespan allows for the FM to access a broader range of knowledge spectrum from private parties, mitigating overfitting on public data, and potentially enabling more generalized and context-aware FMs, while still benefiting from centralized data.

\noindent\textbf{Goal:} 
Enhance FM pre-train methodologies via FL, and allow FMs to foster a deeper understanding of data representations from private data, thereby enhancing the model's capability to generalize across various tasks and domains.

\noindent\textbf{Procedure Overview:} 
As shown in Algorithm~\ref{algo:ffm} and Figure~\ref{fig:FFM_tasks}, FFM pre-training is structured in two phases: centralized pre-training on public data, and federated pre-training on private data. 
% As depicted in Figure~\ref{fig:FFM_tasks}, 
these phases interact via an adaptive switching mechanism, enabling the model to alternate between centralized pre-training (if the centralized public data is available) and federated pre-training. 

% An exemplar FFM pre-training round could commence with pre-training the model on public data for \(n_c\) epochs using existing FM pre-training methods, followed by \(n_f\) epochs on private data via federated learning. This process can also be initiated with federated learning on private data followed by centralized pre-training, rendering a flexible and adaptive pre-training regimen. Existing federated learning algorithms could be tailored for FM pre-training to facilitate the integration of federated learning advantages seamlessly.

\subsection{Federated Foundation Model Fine-tuning}
\label{sec:ft_ffm}
% \subsubsection{Task Definition}
\textbf{Motivation:}
Traditional FM fine-tuning typically involves an offline deployment where the model is fine-tuned on private data, and subsequently isolated. This isolation precludes collaboration among end-users, potentially limiting the FM's efficacy, especially when the local private data is limited and biased.

\noindent\textbf{Goal:}
Leverage the collaborative learning feature of FL, enabling end-users with similar downstream tasks to collaboratively fine-tune FMs while preserving data privacy, thus potentially achieving enhanced performance on downstream tasks. 

\noindent\textbf{Procedure Overview:}
Similar to FFM pre-training, FFM fine-tuning follows the same procedure in Algorithm~\ref{algo:ffm}, FFM fine-tuning builds upon FFM pre-training phase. It employs an adaptive switching mechanism to alternate between centralized fine-tuning on public datasets for benchmark tasks and federated fine-tuning on private data for local tasks. 
As depicted in Figure~\ref{fig:FFM_tasks}, various fine-tuning strategies can be adopted with FFM. These include, but are not limited to, (1) direct fine-tuning of the FM backbone, and (2) Parameter Efficient Fine-tuning (PEFT) of a lightweight adapter head, while keeping the FM backbone frozen.

% Mirroring the FFM pre-training approach, as illustrated in Algorithm~\ref{algo:ffm}, FFM fine-tuning builds upon the pre-training phase. It employs an adaptive switching mechanism to alternate between centralized fine-tuning on public datasets for benchmark tasks, and federated fine-tuning on private data for local tasks. As depicted in Figure~\ref{fig:FFM_tasks}, various fine-tuning strategies can be adopted with FFM. These include, but are not limited to, (1) direct fine-tuning of the FM backbone, and (2) Parameter Efficient Fine-tuning (PEFT) of a lightweight adapter head, while keeping the FM backbone frozen.

\subsection{Federated Prompt Tuning}

% \subsubsection{Task Definition}
\textbf{Motivation:}
Incorporating FL into prompt engineering presents a promising avenue for enhancing the performance of FMs while maintaining data privacy. Specifically, FFMs can assist in utilizing sensitive data for crafting prompt templates and soft prompt tuning, which in turn, enables more accurate and personalized prompt conditioning for tasks.

\noindent\textbf{Goal:}
Collaboratively develop more effective and adaptable prompts without compromising the privacy of sensitive data.

\noindent\textbf{Procedure Overview:}
This subsection primarily explores automated prompt (soft prompt) methods like prompt tuning~\cite{lester2021prompt_tune}, which refines the input prompt to better the model's output. As illustrated in Figure~\ref{fig:FFM_tasks} and the general FFM optimization process in Algorithm~\ref{algo:ffm}, within federated prompt engineering settings, end-users can collaboratively train auto-prompt models (prompt generator components in Figure~\ref{fig:FFM_tasks}) on their local private data and tasks, sharing the learned auto prompt models without disclosing the sensitive data. This collaborative endeavor facilitates the creation of more effective and adaptable prompts, thereby enhancing the overall performance of FMs on downstream tasks.

\subsection{Federated Continual (Lifelong) Learning}
% \subsubsection{Task Definition}
\textbf{Motivation:}
FMs exhibit a significant limitation due to their dependency on pre-trained offline knowledge. For example, ChatGPT's knowledge is up-to-date only until 2021. With the anticipated increase in computational power, FM optimization at the edge may become feasible. FFMs can unlock the possibility of continual and lifelong learning from newly generated private edge data. With its scalability and privacy-preserving nature, FL can harness decentralized power to optimize FMs using emerging private data at the edge, which can serve as a valuable resource for model optimization.
Furthermore, federated continual and lifelong learning could lead to a more efficient utilization of resources. Institutions would no longer necessitate retraining models from scratch with the availability of new data. Through FL, incremental model improvements can be attained, thus diminishing the time and computational resources requisite for model training and refinement.

% Moreover, federated continual/lifelong learning can contribute to more efficient use of resources, as institutions no longer need to retrain their models from scratch whenever new data becomes available. By leveraging FL, institutions can incrementally improve their models, reducing the time and computational resources required for model training and refinement.

\noindent\textbf{Goal:}
Employ FL to harness the computational power at the edge, unlocking the potential for continual and lifelong learning of FMs on newly generated private data at the edge. This approach also aims to keep FMs updated with contemporary knowledge while preserving data privacy.

\noindent\textbf{Procedure Overview:}
As delineated in Sections~\ref{sec:pt_ffm} and \ref{sec:ft_ffm}, establishing an online federated server is essential to facilitate the continuous communication between the server and edge end-users. The FM is updated at the edge based on the newly generated private data and regularly synchronizes with the online server.

\subsection{Federated Retrieval Augmented Generation}
\textbf{Motivation:}
Federated Retrieval Augmented Generation (FRAG) seeks to extend the advantages of Retrieval Augmented Generation (RAG) by leveraging decentralized data across various clients while ensuring privacy preservation. This amalgamation aims to furnish more current and precise responses in a privacy-conducive manner.

% Federated Retrieval Augmented Generation (FRAG) extend the benefits of Retrieval Augmented Generation (RAG) by utilizing decentralized data from various clients while preserving privacy. This integration can lead to more up-to-date and accurate responses in a privacy-preserving manner.

\noindent\textbf{Goal:} 
Integrate FL with the RAG framework to bolster the performance of Language Model Generators (LMGs) in crafting responses, utilizing both centralized and decentralized data sources.

\noindent\textbf{Procedure Overview:}
In the FRAG framework, the procedure unfolds in several distinct phases to ensure both effective data retrieval and privacy preservation. During the retrieval phase, a query is initiated from a user end, which triggers data retrieval from both a centralized server and local databases of clients within a federated network. This query is shared among clients in a privacy-preserving manner, enabling local clients to fetch relevant private data at the edge. 
Following the data retrieval, the generation phase commences where each client independently generates a response based on the retrieved data and the initial query. The responses from all clients are then aggregated in a privacy-preserving manner, ensuring no sensitive information is exposed during the process. Finally, an aggregated response, which encapsulates the collective intelligence of the federated network while preserving user privacy, is relayed back to the user. This structure allows for a more informed and accurate response generation in a decentralized and privacy-preserving environment.

\subsection{Challenges}
\label{sec:pt_challenges}
Despite the benefits associated with FFM, several substantial challenges persist. This subsection enumerates and discusses these general challenges.

\noindent\textbf{Model Size:} 
The substantial size of FMs, such as GPT~\cite{openai2023gpt4} and Llama~\cite{touvron2023llama2}, presents a significant challenge for optimization FMs at the edge, especially when considering the resource-constraint edge devices in FL settings.

\noindent\textbf{Data Quality:} 
The effectiveness of FM pre-training and fine-tuning, including self-supervised pre-training, is heavily contingent on data quality as highlighted in \cite{gunasekar2023textbooks}. Ensuring high-quality data in private federated settings, where data sharing is restricted, presents a notable challenge in filtering out toxic and redundant data.

\noindent\textbf{Computational Cost:}  
Optimizing FMs entails substantial computational cost~\cite{meng2023foundation}. In FL environments, collaborative optimization of FMs at the edge necessitates high hardware specifications for edge clients~\cite{meindl2023cp1,malandrino2021cp2}.

\noindent\textbf{Communication Cost:}  
The routine sharing of model updates, encompassing model weights and gradients, incurs significant communication overhead~\cite{morell2022com_fl,Omair2023com_fl2,mohammadi2021com_fl3,com_fl4} between clients and the server in FL environments.

\noindent\textbf{Data Heterogeneity:} 
In FL, data is often non-identically distributed (non-IID) across clients~\cite{zhao2018noniid,mcmahan2017fedavg}, which could adversely affect the convergence and performance of the optimization process.

\noindent \textbf{Security Attacks:} 
Although FL inherently preserves privacy, ensuring robust privacy guarantees in FFM, especially against sophisticated security attacks, remains vital~\cite{attack1,zhang2022security,liu2022threats}.

\noindent \textbf{Scalability:} 
With the escalating scale of deployment, efficiently managing collaborative training and sharing model updates becomes increasingly challenging~\cite{diaz2023sb1,zawad2022local,kolodziej2021sb2}.

\noindent\textbf{Asynchronous Training:} 
As the number of clients increases, efficiently aggregating updates from a large number of asynchronous clients and ensuring consistent performance scaling is challenging~\cite{wang2022asyncfeded,chen2021afl}.

\noindent\textbf{Non-Stationary Data Distributions:} The perpetually evolving nature of the user data suggests that data distributions may shift over time~\cite{data_distribution}. Ensuring robust model performance amidst such changes is a significant challenge.

\noindent\textbf{Resource Constraints:} The resource-constrained edge devices could impede the optimization process of FMs at the edge.

\noindent\textbf{Global Model Synchronization:} Achieving global model synchronization across all participants while accommodating local updates and ensuring model stability is a nuanced challenge.

\noindent\textbf{Evaluation Metrics:} Establishing robust metrics to evaluate the performance, privacy, and other crucial aspects of the FFM process is pivotal.

\subsection{Other Future Research Directions}
In addition to the potential FFM tasks and general challenges discussed earlier, we outline several potential future research directions below.

\noindent\textbf{Advancement in Edge Hardware:}
Supporting the substantial computational and resource requirements of FM optimization in FL-edge scenarios necessitates significant advancements in edge hardware.

\noindent\textbf{Private-preserve Training Data Process:} 
% The efficacy of self-supervised pre-training is heavily reliant on the quality of data, private preserve training data process methods is essential to FFM, since private data at FL-edge clients are inaccessible, only data owner can access it, and hence it's hard to preprocess and guarantee the data quality at edge. For example, recent work, such as~\cite{gunasekar2023textbooks,li2023textbooks2}, proposes an automatic training data filter to evaluate the data quality, and hence address the data quality issue. 
The success of self-supervised pre-training largely hinges on data quality. In the context of FFM, where private data at FL-edge clients remains inaccessible, and only the data owner can access it, devising private-preserving training data processing methods is crucial. This is to ensure data quality at the edge, where preprocessing is challenging. Recent works, such as~\cite{gunasekar2023textbooks,li2023textbooks2}, propose automatic training data filters to evaluate and enhance data quality, addressing a critical aspect of data processing in FFM.

\noindent\textbf{Collaborative Model Compression:} 
% Reducing the FM size by designing specialized model compression method, such as network pruning and quantization, to help shrink down model size without compromising performance, which is critical for enabling efficient FFM pre-training, especially on edge devices with limited computational resources.
Designing specialized model compression methods, like network pruning and quantization, for heterogeneous-resource edge clients is essential to efficiently utilize the resources at edge clients. It also helps reduce the size of FMs without sacrificing performance. This is particularly critical for environments with limited computational resources.

\noindent\textbf{Neural Architecture Design:} 
% Design computational and hardware efficient neural network architectures.
The design of computational and hardware-efficient neural network architectures is a promising direction to explore, aiming to address the resource constraints and performance requirements in FFM deployment.

\noindent\textbf{Collaborative Self-supervised Learning:} 
% Self-supervised learning dominated FM pre-training, design specialized collaborative self-supervised learning can reasonably utilize decentralized computational power in FL-edge environments.
Self-supervised learning has been a dominant approach for FM pre-training. Developing specialized collaborative self-supervised learning methods can effectively harness decentralized computational power in FL-edge environments.

\noindent\textbf{Collaborative Parameter Efficient Fine-tuning:} 
Designing collaborative parameter-efficient fine-tuning (PEFT) methods is crucial for fine-tuning FMs in FL scenarios, especially given the limited and heterogeneous resource capacities of edge clients.

\noindent\textbf{Robust Model Fusion Algorithms:}
Creating robust algorithms for model fusion is vital to ensure the effective aggregation of model updates from different clients while preserving data privacy and model performance.

\noindent\textbf{Federated Multi-task Learning:}
Exploring federated multi-task learning can facilitate the simultaneous optimization of multiple learning tasks across a federated network, leveraging the collective data and computational resources to improve model performance across various domains.

% 1.Multi-Task and Multi-Modal Learnin
% 2.collaborative PEFT 

% \input{content/05_case_studies}
\section{Conclusion and discussion}

In this paper, we introduced the concept of Federated Foundation Models (FFMs), which integrate Federated Learning (FL) into the lifespan of Foundation Models (FMs). We discussed FFM tasks, general challenges and potential future research directions.
It is important to note that the advancement of computation at edge users is crucial for the widespread adoption of FFMs, and we believe that such advancements will be realized in the near future. As the field of FFM continues to grow, we anticipate the emergence of numerous related research areas, including improved privacy-preserving techniques, the integration of FFM with emerging technologies like IoT and edge computing, and the exploration of FFM in various application domains such as healthcare, finance, and manufacturing.
Additionally, we foresee advancements in adaptive model compression methods for FFM local institutions, communication efficiency research, specialized FL algorithms for efficient updates and aggregation of FFM models, and security attack research. Overall, FFM represents a promising research area in the age of FMs, with the potential to address various challenges in privacy, scalability, and robustness across diverse domains.

% \section{Conclusion}
% Summary of the key points
% Implications of the Federated Foundation Model for Federated Learning research and practice
% Call for further research and exploration

% \maketitleabstract
\section{Acknowledgment}
This research was supported by the National Science Foundation under Grant number 2243775. We also appreciate the provision of computational resources by Intel Labs for this project. Furthermore, we thank the ResearchIT team \footnote{https://researchit.las.iastate.edu} at Iowa State University for their continuous and helpful assistance.

\section{Bibliographical References}\label{sec:reference}

\bibliographystyle{lrec-coling2024-natbib}
\bibliography{reference}

\begin{thebibliography}{51}
\expandafter\ifx\csname natexlab\endcsname\relax\def\natexlab#1{#1}\fi

\bibitem[{Almanifi et~al.(2023)Almanifi, Chow, Tham, Chuah, and Kanesan}]{Omair2023com_fl2}
Omair Rashed~Abdulwareth Almanifi, Chee-Onn Chow, Mau-Luen Tham, Joon~Huang Chuah, and Jeevan Kanesan. 2023.
\newblock \href {https://doi.org/https://doi.org/10.1016/j.iot.2023.100742} {Communication and computation efficiency in federated learning: A survey}.
\newblock \emph{Internet of Things}, 22:100742.

\bibitem[{Brown et~al.(2020)Brown, Mann, Ryder, Subbiah, Kaplan, Dhariwal, Neelakantan, Shyam, Sastry, Askell et~al.}]{brown2020gpt3}
Tom Brown, Benjamin Mann, Nick Ryder, Melanie Subbiah, Jared~D Kaplan, Prafulla Dhariwal, Arvind Neelakantan, Pranav Shyam, Girish Sastry, Amanda Askell, et~al. 2020.
\newblock Language models are few-shot learners.
\newblock \emph{Advances in neural information processing systems}, 33:1877--1901.

\bibitem[{Chen et~al.(2021)Chen, Liao, Hua, Lu, and Yu}]{chen2021afl}
Z~Chen, W~Liao, K~Hua, C~Lu, and W~Yu. 2021.
\newblock Towards asynchronous federated learning for heterogeneous edge-powered internet of things. digit commun netw 7 (3): 317--326.

\bibitem[{Deng et~al.(2020)Deng, Kamani, and Mahdavi}]{deng2020adaptive_pfl}
Yuyang Deng, Mohammad~Mahdi Kamani, and Mehrdad Mahdavi. 2020.
\newblock Adaptive personalized federated learning.
\newblock \emph{arXiv preprint arXiv:2003.13461}.

\bibitem[{D{\'\i}az and Garc{\'\i}a(2023)}]{diaz2023sb1}
Judith S{\'a}inz-Pardo D{\'\i}az and {\'A}lvaro~L{\'o}pez Garc{\'\i}a. 2023.
\newblock Study of the performance and scalability of federated learning for medical imaging with intermittent clients.
\newblock \emph{Neurocomputing}, 518:142--154.

\bibitem[{Dosovitskiy et~al.(2020)Dosovitskiy, Beyer, Kolesnikov, Weissenborn, Zhai, Unterthiner, Dehghani, Minderer, Heigold, Gelly et~al.}]{dosovitskiy2020vit}
Alexey Dosovitskiy, Lucas Beyer, Alexander Kolesnikov, Dirk Weissenborn, Xiaohua Zhai, Thomas Unterthiner, Mostafa Dehghani, Matthias Minderer, Georg Heigold, Sylvain Gelly, et~al. 2020.
\newblock An image is worth 16x16 words: Transformers for image recognition at scale.
\newblock \emph{arXiv preprint arXiv:2010.11929}.

\bibitem[{Gunasekar et~al.(2023)Gunasekar, Zhang, Aneja, Mendes, Del~Giorno, Gopi, Javaheripi, Kauffmann, de~Rosa, Saarikivi et~al.}]{gunasekar2023textbooks}
Suriya Gunasekar, Yi~Zhang, Jyoti Aneja, Caio C{\'e}sar~Teodoro Mendes, Allie Del~Giorno, Sivakanth Gopi, Mojan Javaheripi, Piero Kauffmann, Gustavo de~Rosa, Olli Saarikivi, et~al. 2023.
\newblock Textbooks are all you need.
\newblock \emph{arXiv preprint arXiv:2306.11644}.

\bibitem[{Han et~al.(2022)Han, Zhao, Ding, Liu, and Sun}]{han2022ptr_prompt_tune}
Xu~Han, Weilin Zhao, Ning Ding, Zhiyuan Liu, and Maosong Sun. 2022.
\newblock Ptr: Prompt tuning with rules for text classification.
\newblock \emph{AI Open}, 3:182--192.

\bibitem[{Hu et~al.(2022)Hu, yelong shen, Wallis, Allen-Zhu, Li, Wang, Wang, and Chen}]{hu2022lora}
Edward~J Hu, yelong shen, Phillip Wallis, Zeyuan Allen-Zhu, Yuanzhi Li, Shean Wang, Lu~Wang, and Weizhu Chen. 2022.
\newblock \href {https://openreview.net/forum?id=nZeVKeeFYf9} {Lo{RA}: Low-rank adaptation of large language models}.
\newblock In \emph{International Conference on Learning Representations}.

\bibitem[{Jiang et~al.(2022)Jiang, Wang, Valls, Ko, Lee, Leung, and Tassiulas}]{jiang2022pruneFL}
Yuang Jiang, Shiqiang Wang, Victor Valls, Bong~Jun Ko, Wei-Han Lee, Kin~K Leung, and Leandros Tassiulas. 2022.
\newblock Model pruning enables efficient federated learning on edge devices.
\newblock \emph{IEEE Transactions on Neural Networks and Learning Systems}.

\bibitem[{Karimireddy et~al.(2020)Karimireddy, Kale, Mohri, Reddi, Stich, and Suresh}]{karimireddy2020scaffold}
Sai~Praneeth Karimireddy, Satyen Kale, Mehryar Mohri, Sashank Reddi, Sebastian Stich, and Ananda~Theertha Suresh. 2020.
\newblock Scaffold: Stochastic controlled averaging for federated learning.
\newblock In \emph{International Conference on Machine Learning}, pages 5132--5143. PMLR.

\bibitem[{Kenton and Toutanova(2019)}]{kenton2019bert}
Jacob Devlin Ming-Wei~Chang Kenton and Lee~Kristina Toutanova. 2019.
\newblock Bert: Pre-training of deep bidirectional transformers for language understanding.
\newblock In \emph{Proceedings of naacL-HLT}, volume~1, page~2.

\bibitem[{Ko{\l}odziej and Ro{\'s}ciszewski(2021)}]{kolodziej2021sb2}
Tomasz Ko{\l}odziej and Pawe{\l} Ro{\'s}ciszewski. 2021.
\newblock Towards scalable simulation of federated learning.
\newblock In \emph{Neural Information Processing: 28th International Conference, ICONIP 2021, Sanur, Bali, Indonesia, December 8--12, 2021, Proceedings, Part V 28}, pages 248--256. Springer.

\bibitem[{Lester et~al.(2021)Lester, Al-Rfou, and Constant}]{lester2021prompt_tune}
Brian Lester, Rami Al-Rfou, and Noah Constant. 2021.
\newblock The power of scale for parameter-efficient prompt tuning.
\newblock \emph{arXiv preprint arXiv:2104.08691}.

\bibitem[{Li et~al.(2023)Li, Bubeck, Eldan, Del~Giorno, Gunasekar, and Lee}]{li2023textbooks2}
Yuanzhi Li, S{\'e}bastien Bubeck, Ronen Eldan, Allie Del~Giorno, Suriya Gunasekar, and Yin~Tat Lee. 2023.
\newblock Textbooks are all you need ii: phi-1.5 technical report.
\newblock \emph{arXiv preprint arXiv:2309.05463}.

\bibitem[{Lin et~al.(2020)Lin, Kong, Stich, and Jaggi}]{lin2020feddf}
Tao Lin, Lingjing Kong, Sebastian~U Stich, and Martin Jaggi. 2020.
\newblock Ensemble distillation for robust model fusion in federated learning.
\newblock \emph{Advances in Neural Information Processing Systems}, 33:2351--2363.

\bibitem[{Liu et~al.(2021)Liu, Shen, Zhang, Dolan, Carin, and Chen}]{liu2021icl_makes}
Jiachang Liu, Dinghan Shen, Yizhe Zhang, Bill Dolan, Lawrence Carin, and Weizhu Chen. 2021.
\newblock What makes good in-context examples for gpt-$3 $?
\newblock \emph{arXiv preprint arXiv:2101.06804}.

\bibitem[{Liu et~al.(2022)Liu, Xu, and Wang}]{liu2022threats}
Pengrui Liu, Xiangrui Xu, and Wei Wang. 2022.
\newblock Threats, attacks and defenses to federated learning: issues, taxonomy and perspectives.
\newblock \emph{Cybersecurity}, 5(1):1--19.

\bibitem[{Liu et~al.(2020)Liu, Chen, Chen, and Zhang}]{liu2020momentum_fl}
Wei Liu, Li~Chen, Yunfei Chen, and Wenyi Zhang. 2020.
\newblock Accelerating federated learning via momentum gradient descent.
\newblock \emph{IEEE Transactions on Parallel and Distributed Systems}, 31(8):1754--1766.

\bibitem[{Lyu et~al.(2022)Lyu, Yu, Ma, Chen, Sun, Zhao, Yang, and Yu}]{attack1}
Lingjuan Lyu, Han Yu, Xingjun Ma, Chen Chen, Lichao Sun, Jun Zhao, Qiang Yang, and Philip~S. Yu. 2022.
\newblock \href {https://doi.org/10.1109/TNNLS.2022.3216981} {Privacy and robustness in federated learning: Attacks and defenses}.
\newblock \emph{IEEE Transactions on Neural Networks and Learning Systems}, pages 1--21.

\bibitem[{Malandrino and Chiasserini(2021)}]{malandrino2021cp2}
Francesco Malandrino and Carla~Fabiana Chiasserini. 2021.
\newblock Toward node liability in federated learning: Computational cost and network overhead.
\newblock \emph{IEEE Communications Magazine}, 59(9):72--77.

\bibitem[{McMahan et~al.(2017)McMahan, Moore, Ramage, Hampson, and y~Arcas}]{mcmahan2017fedavg}
Brendan McMahan, Eider Moore, Daniel Ramage, Seth Hampson, and Blaise~Aguera y~Arcas. 2017.
\newblock Communication-efficient learning of deep networks from decentralized data.
\newblock In \emph{Artificial intelligence and statistics}, pages 1273--1282. PMLR.

\bibitem[{Meindl and Moser(2023)}]{meindl2023cp1}
Rainer Meindl and Bernhard~A Moser. 2023.
\newblock Measuring overhead costs of federated learning systems by eavesdropping.
\newblock In \emph{International Conference on Database and Expert Systems Applications}, pages 33--42. Springer.

\bibitem[{Meng et~al.(2023)Meng, Shao, Peng, Jiang, Zhang, Qiao, and Luo}]{meng2023foundation}
Fanqing Meng, Wenqi Shao, Zhanglin Peng, Chonghe Jiang, Kaipeng Zhang, Yu~Qiao, and Ping Luo. 2023.
\newblock \href {http://arxiv.org/abs/2308.06262} {Foundation model is efficient multimodal multitask model selector}.

\bibitem[{Min et~al.(2021)Min, Lewis, Zettlemoyer, and Hajishirzi}]{min2021icl_metaicl}
Sewon Min, Mike Lewis, Luke Zettlemoyer, and Hannaneh Hajishirzi. 2021.
\newblock Metaicl: Learning to learn in context.
\newblock \emph{arXiv preprint arXiv:2110.15943}.

\bibitem[{Min et~al.(2022)Min, Lyu, Holtzman, Artetxe, Lewis, Hajishirzi, and Zettlemoyer}]{min2022icl_rethinking}
Sewon Min, Xinxi Lyu, Ari Holtzman, Mikel Artetxe, Mike Lewis, Hannaneh Hajishirzi, and Luke Zettlemoyer. 2022.
\newblock Rethinking the role of demonstrations: What makes in-context learning work?
\newblock \emph{arXiv preprint arXiv:2202.12837}.

\bibitem[{Mohammadi et~al.(2021)Mohammadi, Bai, Fan, Song, Yi, and Liu}]{mohammadi2021com_fl3}
Nima Mohammadi, Jianan Bai, Qiang Fan, Yifei Song, Yang Yi, and Lingjia Liu. 2021.
\newblock \href {http://arxiv.org/abs/2101.12240} {Differential privacy meets federated learning under communication constraints}.

\bibitem[{Muñoz et~al.(2024{\natexlab{a}})Muñoz, Yuan, and Jain}]{shears2024}
J.~Pablo Muñoz, Jinjie Yuan, and Nilesh Jain. 2024{\natexlab{a}}.
\newblock Shears: Unstructured sparsity with neural low-rank adapter search.

\bibitem[{Muñoz et~al.(2024{\natexlab{b}})Muñoz, Yuan, Zheng, and Jain}]{lonas2024}
J.~Pablo Muñoz, Jinjie Yuan, Yi~Zheng, and Nilesh Jain. 2024{\natexlab{b}}.
\newblock Lonas: Elastic low-rank adapters for efficient large language models.
\newblock In \emph{The 2024 Joint International Conference on Computational Linguistics, Language Resources and Evaluation}.

\bibitem[{Nguyen et~al.(2023)Nguyen, Yu, Mu\~{n}oz, and Jannesari}]{nguyen2024flkd}
Duy~Phuong Nguyen, Sixing Yu, J.~Pablo Mu\~{n}oz, and Ali Jannesari. 2023.
\newblock \href {https://doi.org/10.1145/3624062.3626325} {Enhancing heterogeneous federated learning with knowledge extraction and multi-model fusion}.
\newblock In \emph{Proceedings of the SC '23 Workshops of The International Conference on High Performance Computing, Network, Storage, and Analysis}, SC-W '23, page 36–43, New York, NY, USA. Association for Computing Machinery.

\bibitem[{OpenAI(2023)}]{openai2023gpt4}
OpenAI. 2023.
\newblock \href {http://arxiv.org/abs/2303.08774} {Gpt-4 technical report}.

\bibitem[{Radford et~al.(2021)Radford, Kim, Hallacy, Ramesh, Goh, Agarwal, Sastry, Askell, Mishkin, Clark et~al.}]{radford2021clip}
Alec Radford, Jong~Wook Kim, Chris Hallacy, Aditya Ramesh, Gabriel Goh, Sandhini Agarwal, Girish Sastry, Amanda Askell, Pamela Mishkin, Jack Clark, et~al. 2021.
\newblock Learning transferable visual models from natural language supervision.
\newblock In \emph{International conference on machine learning}, pages 8748--8763. PMLR.

\bibitem[{Radford et~al.(2019)Radford, Wu, Child, Luan, Amodei, Sutskever et~al.}]{radford2019gpt2}
Alec Radford, Jeffrey Wu, Rewon Child, David Luan, Dario Amodei, Ilya Sutskever, et~al. 2019.
\newblock Language models are unsupervised multitask learners.
\newblock \emph{OpenAI blog}, 1(8):9.

\bibitem[{Rubin et~al.(2021)Rubin, Herzig, and Berant}]{rubin2021icl_retrive_prompt}
Ohad Rubin, Jonathan Herzig, and Jonathan Berant. 2021.
\newblock Learning to retrieve prompts for in-context learning.
\newblock \emph{arXiv preprint arXiv:2112.08633}.

\bibitem[{Sanh et~al.(2021)Sanh, Webson, Raffel, Bach, Sutawika, Alyafeai, Chaffin, Stiegler, Scao, Raja et~al.}]{sanh2021multitask_automated_prompt}
Victor Sanh, Albert Webson, Colin Raffel, Stephen~H Bach, Lintang Sutawika, Zaid Alyafeai, Antoine Chaffin, Arnaud Stiegler, Teven~Le Scao, Arun Raja, et~al. 2021.
\newblock Multitask prompted training enables zero-shot task generalization.
\newblock \emph{arXiv preprint arXiv:2110.08207}.

\bibitem[{Tan et~al.(2022)Tan, Yu, Cui, and Yang}]{tan2022personalizedfl}
Alysa~Ziying Tan, Han Yu, Lizhen Cui, and Qiang Yang. 2022.
\newblock Towards personalized federated learning.
\newblock \emph{IEEE Transactions on Neural Networks and Learning Systems}.

\bibitem[{Touvron et~al.(2023{\natexlab{a}})Touvron, Lavril, Izacard, Martinet, Lachaux, Lacroix, Rozi{\`e}re, Goyal, Hambro, Azhar et~al.}]{touvron2023llama}
Hugo Touvron, Thibaut Lavril, Gautier Izacard, Xavier Martinet, Marie-Anne Lachaux, Timoth{\'e}e Lacroix, Baptiste Rozi{\`e}re, Naman Goyal, Eric Hambro, Faisal Azhar, et~al. 2023{\natexlab{a}}.
\newblock Llama: Open and efficient foundation language models.
\newblock \emph{arXiv preprint arXiv:2302.13971}.

\bibitem[{Touvron et~al.(2023{\natexlab{b}})Touvron, Martin, Stone, Albert, Almahairi, Babaei, Bashlykov, Batra, Bhargava, Bhosale et~al.}]{touvron2023llama2}
Hugo Touvron, Louis Martin, Kevin Stone, Peter Albert, Amjad Almahairi, Yasmine Babaei, Nikolay Bashlykov, Soumya Batra, Prajjwal Bhargava, Shruti Bhosale, et~al. 2023{\natexlab{b}}.
\newblock Llama 2: Open foundation and fine-tuned chat models.
\newblock \emph{arXiv preprint arXiv:2307.09288}.

\bibitem[{WANG et~al.(2019)WANG, WANG, and LI}]{com_fl4}
Luping WANG, Wei WANG, and Bo~LI. 2019.
\newblock \href {https://doi.org/10.1109/ICDCS.2019.00099} {Cmfl: Mitigating communication overhead for federated learning}.
\newblock In \emph{2019 IEEE 39th International Conference on Distributed Computing Systems (ICDCS)}, pages 954--964.

\bibitem[{Wang et~al.(2022)Wang, Yang, He, Shi, and Chen}]{wang2022asyncfeded}
Qiyuan Wang, Qianqian Yang, Shibo He, Zhiguo Shi, and Jiming Chen. 2022.
\newblock Asyncfeded: Asynchronous federated learning with euclidean distance based adaptive weight aggregation.
\newblock \emph{arXiv preprint arXiv:2205.13797}.

\bibitem[{Wei et~al.(2021)Wei, Bosma, Zhao, Guu, Yu, Lester, Du, Dai, and Le}]{wei2021prompt}
Jason Wei, Maarten Bosma, Vincent~Y Zhao, Kelvin Guu, Adams~Wei Yu, Brian Lester, Nan Du, Andrew~M Dai, and Quoc~V Le. 2021.
\newblock Finetuned language models are zero-shot learners.
\newblock \emph{arXiv preprint arXiv:2109.01652}.

\bibitem[{Yu et~al.(2022{\natexlab{a}})Yu, Nguyen, Abebe, Qian, Anwar, and Jannesari}]{yu2022spatl}
Sixing Yu, Phuong Nguyen, Waqwoya Abebe, Wei Qian, Ali Anwar, and Ali Jannesari. 2022{\natexlab{a}}.
\newblock Spatl: salient parameter aggregation and transfer learning for heterogeneous federated learning.
\newblock In \emph{2022 SC22: International Conference for High Performance Computing, Networking, Storage and Analysis (SC)}, pages 495--508. IEEE Computer Society.

\bibitem[{Yu et~al.(2022{\natexlab{b}})Yu, Nguyen, Abebe, Stanley, Muñoz, and Jannesari}]{yu2022rafl}
Sixing Yu, Phuong Nguyen, Waqwoya Abebe, Justin Stanley, Pablo Muñoz, and Ali Jannesari. 2022{\natexlab{b}}.
\newblock Resource-aware heterogeneous federated learning using neural architecture search.
\newblock \emph{arXiv preprint arXiv:2211.05716}.

\bibitem[{Yu et~al.(2021)Yu, Nguyen, Anwar, and Jannesari}]{yu2021feddp}
Sixing Yu, Phuong Nguyen, Ali Anwar, and Ali Jannesari. 2021.
\newblock Adaptive dynamic pruning for non-iid federated learning.
\newblock \emph{arXiv preprint arXiv:2106.06921}.

\bibitem[{Yu et~al.(2022{\natexlab{c}})Yu, Qian, and Jannesari}]{yu2022kdfl}
Sixing Yu, Wei Qian, and Ali Jannesari. 2022{\natexlab{c}}.
\newblock Resource-aware federated learning using knowledge extraction and multi-model fusion.
\newblock \emph{arXiv preprint arXiv:2208.07978}.

\bibitem[{Zawad et~al.(2022)Zawad, Yan, and Anwar}]{zawad2022local}
Syed Zawad, Feng Yan, and Ali Anwar. 2022.
\newblock Local training and scalability of federated learning systems.
\newblock In \emph{Federated Learning: A Comprehensive Overview of Methods and Applications}, pages 213--233. Springer.

\bibitem[{Zhang et~al.(2022{\natexlab{a}})Zhang, Tao, Shi, and Bi}]{data_distribution}
Hongwei Zhang, Meixia Tao, Yuanming Shi, and Xiaoyan Bi. 2022{\natexlab{a}}.
\newblock \href {https://doi.org/10.1109/ICC45855.2022.9838703} {Federated multi-task learning with non-stationary heterogeneous data}.
\newblock In \emph{ICC 2022 - IEEE International Conference on Communications}, pages 4950--4955.

\bibitem[{Zhang et~al.(2022{\natexlab{b}})Zhang, Zhu, Wang, Zhao, Xu, Li et~al.}]{zhang2022security}
Junpeng Zhang, Hui Zhu, Fengwei Wang, Jiaqi Zhao, Qi~Xu, Hui Li, et~al. 2022{\natexlab{b}}.
\newblock Security and privacy threats to federated learning: Issues, methods, and challenges.
\newblock \emph{Security and Communication Networks}, 2022.

\bibitem[{Zhao et~al.(2018)Zhao, Li, Lai, Suda, Civin, and Chandra}]{zhao2018noniid}
Yue Zhao, Meng Li, Liangzhen Lai, Naveen Suda, Damon Civin, and Vikas Chandra. 2018.
\newblock Federated learning with non-iid data.
\newblock \emph{arXiv preprint arXiv:1806.00582}.

\bibitem[{Zhou et~al.(2022)Zhou, Yang, Loy, and Liu}]{zhou2022automated_prompt}
Kaiyang Zhou, Jingkang Yang, Chen~Change Loy, and Ziwei Liu. 2022.
\newblock Learning to prompt for vision-language models.
\newblock \emph{International Journal of Computer Vision}, 130(9):2337--2348.

\bibitem[{Ángel Morell et~al.(2022)Ángel Morell, Dahi, Chicano, Luque, and Alba}]{morell2022com_fl}
José Ángel Morell, Zakaria~Abdelmoiz Dahi, Francisco Chicano, Gabriel Luque, and Enrique Alba. 2022.
\newblock \href {http://arxiv.org/abs/2204.02183} {Optimising communication overhead in federated learning using nsga-ii}.

\end{thebibliography}

% \section{Language Resource References}
% \label{lr:ref}
% \bibliographystylelanguageresource{lrec-coling2024-natbib}
% \bibliographylanguageresource{languageresource}

\end{document}